\documentclass[letterpaper]{article} 
\usepackage{aaai24}  
\usepackage{amsfonts}
\usepackage{amsmath}
\usepackage{amsthm}
\usepackage[bb=dsserif]{mathalpha}
\usepackage{bm}
\newtheorem{theorem}{\textbf{Theorem}}

\newtheorem{lemma}{\textbf{Lemma}}

\usepackage{times}  
\usepackage{helvet}  
\usepackage{courier}  
\usepackage[hyphens]{url}  
\usepackage{graphicx} 
\usepackage{natbib}  
\usepackage{caption} 
\usepackage{algorithm}
\usepackage{algorithmic}
\usepackage{color}

\usepackage{cleveref}
\usepackage{newfloat}
\usepackage{listings}
\DeclareCaptionStyle{ruled}{labelfont=normalfont,labelsep=colon,strut=off} 
\floatstyle{ruled}
\newfloat{listing}{tb}{lst}{}
\floatname{listing}{Listing}
\title{Large-Scale Non-convex Stochastic Constrained Distributionally Robust Optimization}
\author{
    Qi Zhang \textsuperscript{\rm 1}, Yi Zhou \textsuperscript{\rm 2}, Ashley Prater-Bennette \textsuperscript{\rm 3}, Lixin Shen \textsuperscript{\rm 4}, Shaofeng Zou \textsuperscript{\rm 1}
}
\affiliations{
    \textsuperscript{\rm 1}University at Buffalo\\
    \textsuperscript{\rm 2}The University of Utah\\
    \textsuperscript{\rm 3}Air Force Research Laboratory\\
    \textsuperscript{\rm 4}Syracuse University\\
    qzhang48@buffalo.edu, yi.zhou@utah.edu, ashley.prater-bennette@us.af.mil, lshen03@syr.edu, szou3@buffalo.edu

}

\usepackage{bibentry}

\begin{document}

    \maketitle
\begin{abstract}
Distributionally robust optimization (DRO) is a powerful framework for training robust models against data distribution shifts. This paper focuses on constrained DRO, which has an explicit characterization of the robustness level. Existing studies on constrained DRO mostly focus on convex loss function, and exclude the practical and challenging case with non-convex loss function, e.g., neural network. This paper develops a  stochastic algorithm and its performance analysis for non-convex constrained DRO. The computational complexity of our stochastic algorithm at each iteration is independent of the overall dataset size, and thus is suitable for large-scale applications. We focus on the general Cressie-Read family divergence defined uncertainty set which includes $\chi^2$-divergences as a special case. {
We prove that our algorithm finds an $\epsilon$-stationary point with a computational complexity of $\mathcal O(\epsilon^{-3k_*-5})$, where $k_*$ is the parameter of the Cressie-Read divergence. The numerical results indicate that our method outperforms existing methods.} Our method also applies to the smoothed conditional value at risk (CVaR) DRO. 
\end{abstract}

\noindent
\section{Introduction}
Machine learning algorithms typically employ the approach of  Empirical Risk Minimization (ERM), which minimizes the expected loss under the empirical distribution $P_0$ of the training dataset and assumes that test samples are generated from the same distribution. However, in practice, there usually exists a mismatch between the training and testing distributions due to various reasons, for example, in domain adaptation tasks domains differ from training to testing \cite{blitzer2006domain,daume2006domain}; test samples were selected from minority groups which are underrepresented in the training dataset \cite{grother2011report,hovy2015tagging} and  there might exist potential adversarial attacks \cite{goodfellow2014explaining,madry2017towards}. Such a mismatch may lead to a significant performance degradation.

This challenge spurred noteworthy efforts on developing
 a framework of Distributionally Robust Optimization (DRO) e.g.,  \cite{ben2013robust,shapiro2017distributionally,rahimian2019distributionally}. Rather than minimizing the expected loss under one fixed distribution,  in DRO, one seeks to optimize the expected loss under the worst-case distribution in an uncertainty set of distributions.


Specifically, DRO aims to solve the following problem:
\begin{flalign}\label{eq:dro}
    \inf_x \sup_{Q\sim \mathcal U(P_0)} \mathbb E_{S\sim Q}  \left[\ell(x;S)\right],
\end{flalign}
where $\mathcal U(P_0)$ is an uncertainty set of distributions 
centered at $P_0$, $P_0$ is the empirical distribution of the training dataset, $\ell$ is the loss function, and $x$ is 
the optimization variable. For example, the uncertainty set can be defined as 
\begin{align}
    \mathcal U(P_0):=\{Q:D(Q\|P_0)\leq \rho\},
\end{align}
where $D$ is some distance-like metric, e.g., Kullback-Leibler (KL) divergence and $\chi^2$ divergence, and $\rho$ is the uncertainty level.
In practice, for ease of implementation and analysis, a relaxed formulation of  \cref{eq:dro}, which is referred to as the penalized DRO,  is usually solved \cite{levy2020large,jin2021non,qi2021online,sinha2017certifying}: 
\begin{flalign}\label{eq:dro1}
     \inf_x \sup_{Q} \mathbb E_{S\sim Q} \ \left[\ell(x;S)\right]-\lambda D(Q\|P_0),
\end{flalign}
where $\lambda>0$ is a fixed hyperparameter that needs to be chosen manually.
In contrast to constrained DRO in eq. \eqref{eq:dro}, a regularization term is added to the objective function to  keep  the distribution $Q$ and the  distribution $P_0$ close, and the hyperparameter $\lambda$ is manually  chosen beforehand to control the tradeoff with minimizing the loss. 
 Compared with the penalized DRO setting, the constrained DRO problem in \cref{eq:dro}  requires that the distribution $Q$ to be strictly in the uncertainty set, and searches for the optimal solution under the worst-case distribution in the uncertainty set. Therefore, the obtained solution from the constrained DRO is minimax optimal for distributions in the uncertainty set, whereas it is hard to get such a guarantee for the penalized DRO relaxation. In this paper, we focus on the challenging constrained DRO problem in \cref{eq:dro}.


Existing studies on constrained DRO are limited to convex loss functions or require some additional assumptions  \cite{soma2020statistical,hashimoto2018fairness,levy2020large,duchi2018learning,duchi2021statistics,qi2022stochastic,wang2021sinkhorn}. Little understanding on the practical non-convex loss functions, e.g., neural network, is known. In this paper, we focus on the constrained DRO problem with non-convex loss. 

DRO problems under different uncertainty sets are fundamentally different. As will be discussed later in related works, there is  a rich literature on DRO with various uncertainty sets. In this paper, we focus on the general Cressie-Read family divergence defined uncertainty set \cite{duchi2018learning,jin2021non}, which includes, e.g., $\chi^2$ divergence, as  a special case (see \Cref{sec:pre} for more details). We also investigate the smoothed conditional value at risk (CVaR) DRO problem.

More importantly, we focus on the practical yet challenging large-scale scenario, where $P_0$ is the empirical distribution of $N$ samples and $N$ is very large. 
In classic stochastic optimization problems, e.g., ERM, it is easy to get an unbiased estimate of the gradient using only a few samples, and therefore the computational complexity at each iteration is independent of the training dataset size. However, in the DRO problems, due to taking the worst-case distributions in the objective, the problem becomes challenging. Many existing DRO algorithms incur a complexity that increases linearly (or even worse) in the   training dataset size \cite{duchi2018learning,namkoong2016stochastic,ghosh2018efficient}, which is not feasible for large-scale applications.  
In this paper, we will design a stochastic algorithm with  computational complexity at each iteration being independent of the training dataset  size \cite{qi2022stochastic,wang2021sinkhorn,levy2020large}.


\subsection{Challenges and Contributions}
The key challenges and main
contributions in this paper are summarized as follows.
\begin{itemize}
    \item For large-scale applications, the number of training samples is large, and therefore directly computing the full gradient is not practical. Nevertheless, as discussed above, it is challenging to obtain an unbiased estimate of the gradient for DRO problems using only a few samples. 
    For $\varphi$-divergence DRO problem, the distributions in the uncertainty set are continuous w.r.t. the training distribution. Thus the distributions in the uncertainty set can be parameterized by an $N$-dimensional vector \cite{namkoong2016stochastic}.  Then the DRO problem becomes a min-max problem and primal-dual algorithms \cite{rafique2022weakly,lin2020gradient,xu2023unified} can be used directly. 
    Subsampling methods in DRO were also studied in \cite{namkoong2016stochastic,ghosh2018efficient}.
    However, all the above studies require a computational complexity  linear or even worse in the training dataset size at each iteration and thus is prohibitive in
    large-scale applications. In \cite{levy2020large}, an efficient subsampling method was proposed, where the batch size is independent of the training dataset size. However, they only showed the sampling bias for $\chi^2$ and CVaR DRO problems. In this paper, we generalize the analysis of the bias in \cite{levy2020large} to the general Cressie-Read family. We further develop a Frank-Wolfe update on the dual variables in order to bound the gap between the objective and its optimal value given the optimization variable $x$ and the biased estimate.
    \item The second challenge is due to the  non-convex loss function. Existing studies for the Cressie-Read divergence family \cite{duchi2018learning,levy2020large} are limited to the case with convex loss function, and their approach does not generalize to the non-convex case. The key difficulty lies in that the subgradient of the objective function can not be obtained via subdifferential for non-convex loss functions.
Instead of explicitly calculating the worst-case distribution as in \cite{duchi2018learning,levy2020large}, we propose to design an algorithm for the dual problem which optimizes the objective under a known distribution. Thus the gradient of the objective can be efficiently obtained.

\item The third challenge is that the dual form of constrained DRO is neither smooth nor Lipschitz, making the convergence analysis difficult. 
Existing studies, e.g., \cite{wang2021sinkhorn}, assume that the optimal dual variable is bounded away from zero, i.e.,  $\lambda^*>\lambda_0$ for some $\lambda_0>0$, so that it is sufficient to consider $\lambda\geq\lambda_0$. However, this assumption may not necessarily be true as shown in \cite{wang2021sinkhorn,hu2013kullback}. In this paper, we generalize the idea in \cite{qi2022stochastic,levy2020large} to the general Cressie-Read divergence family. We design an approximation of the original problem, and show that it is smooth and Lipschitz. The approximation error can be made arbitrarily small so that the solution to the approximation is still a good solution to the original. 
We add a regularizer to the objective and at the same time we keep the hard constraint. In this way, we can guarantee that its dual variable $\lambda$ has a positive lower bound. 
\item We design a novel algorithm to solve the approximated problem and prove it converges to a stationary point of the constrained DRO problem.
{
The numerical results show that our proposed algorithm outperforms existing methods. 
}

\end{itemize}
\subsection{Related Work}
\subsubsection{Various Uncertainty Sets.}
$\varphi$-divergence DRO problems \cite{ali1966general,csiszar1967information} were widely studied, for example, CVaR in \cite{rockafellar2000optimization,soma2020statistical,curi2020adaptive,tamar2015optimizing},  $\chi^2$-divergence in \cite{hashimoto2018fairness,ghosh2018efficient,levy2020large},  KL-divergence in \cite{qi2021online,qi2022stochastic,hu2013kullback} and  Sinkhorn distance \cite{wang2021sinkhorn}, a variant of Wasserstein distance based on entropic regularization. 
However,  the above studies are for some specific divergence function and can not be extended directly to the general Cressie-Read divergence family.
\subsubsection{Penalized DRO.}
The general $\varphi$-divergence DRO problem was studied in \cite{jin2021non} where the proposed algorithm works for any divergence function with a smooth conjugate. The authors also designed a smoothed version of the CVaR problem and showed their method converges to a stationary point. However, their method is for the penalized formulation and does not generalize to the constrained DRO. In this paper, we focus on  the challenging constrained  DRO,   the solution of which is minimax optimal over the uncertainty set. Our proposed algorithm can also be applied to solve the smoothed   CVaR problem in the constrained setting.
\subsubsection{Constrained DRO with Convex Loss.}
The general $\varphi$-divergence constrained DRO problem was studied in \cite{namkoong2016stochastic}. Instead of optimizing from the dual form, the authors treat the worst-case distribution as a $N$-dimentional vector and implement a stochastic primal-dual method to solve the min-max problem. However, the computational complexity at each iteration is linear in the number of the training samples and can not be used in large-scale applications. The same problem was further studied in \cite{duchi2021statistics}. The authors pointed out that minimizing constrained DRO with $\varphi$-divergence is equivalent to adding
variance regularization for the Empirical Risk Minimization (ERM) objective. The general
Cressie-Read divergence family DRO problem was studied in  \cite{duchi2018learning}, where the basic idea is to calculate the worst-case distribution for the constrained DRO first and then use the subdifferential to get the subgradient. Furthermore,   the $\chi^2$ and CVaR DRO problems were studied in \cite{levy2020large}. Compared with the method in \cite{duchi2018learning}, they calculate the worst-case distribution for the penalized DRO and then optimize both the Lagrange multiplier and the loss function. This approach converges to the optimal solution with a reduced complexity. Their method can be extended to the general Cressie-Read divergence family. However, all the above papers are limited to the case with convex loss function. To the best of our knowledge, our work is the first paper on  large-scale non-convex constrained DRO with the general Cressie-Read divergence family. We note that the KL DRO was studied in \cite{qi2022stochastic}, which however needs an exponential computational complexity. We achieve a polynomial computational complexity for the Cressie-Read divergence family.

\noindent 
\section{Preliminaries and Problem Model}\label{sec:pre}
\subsection{Notations} 
Let $s$ be a sample in $\mathbb S$ and $P_0$ be the distribution on the points $\{s_i\}_{i=1}^N$, where $N$ is the size of the support. Denote by $\Delta^{N}:=\{\mathbf p\in \mathbb R^n|\sum_{i=1}^{N} p_i=1, p_i\ge 0\}$ the ${N}$-dimensional probability simplex. Denote by $x\in \mathbb R^d$ the optimization variable. We denote by $\mathbb{1}_{\mathbb X}(x)$  the indicator function, where $\mathbb{1}_{\mathbb X}(x)=0$ if $x\in \mathbb X$, otherwise $\mathbb{1}_{\mathbb X}(x)=\infty$.
Let $\ell:\mathbb R^d \times \mathbb S\to \mathbb R$ be a non-convex loss function. Let $\|\cdot\|$ be the Euclidean norm and $(t)_+:=\max\{t,0\}$ be the positive part of $t\in \mathbb R$.
Denote $\nabla_x$ by the gradient to $x$.
For a function $f:\mathbb R^d\to \mathbb R,$ a point $x\in \mathbb R^d$ is said to be an $\epsilon$-optimal solution if $|f(x)-f(x^*)|\le \epsilon$, where $f(x^*)$ is the optimal value of $f$. If the function $f$ is differentiable, a point $x\in \mathbb R^d$ is said to be first-order $\epsilon$-stationary if $\|\nabla f(x)\|\le \epsilon$.

\subsection{Assumptions}
In this paper, we take the following  standard assumptions that are commonly used in the DRO literature \cite{duchi2018learning,levy2020large,qi2021online,qi2022stochastic,wang2021sinkhorn,soma2020statistical}:
\begin{itemize}
    \item The non-convex loss
function is bounded: $0\le \ell(x;s)\le B$ for some $B>0$, $\forall x\in \mathbb R^d, s\in \mathbb S$.
\item The non-convex loss
function is $G$-Lipschitz such that $|\ell(x_1;s)-\ell(x_2;s)|\le G\|x_1-x_2\|$ and $L$-smooth such that $\|\nabla_x \ell(x_1;s)-\nabla_x \ell(x_2;s)\|\le L\|x_1-x_2\|$ for any $x_1,x_2 \in \mathbb R^d$ and $s\in \mathbb S$.
\end{itemize}
\subsection{DRO Objective and Its Dual Form}
In ERM, the goal is to solve 
\begin{flalign}
    \inf_x \mathbb E_{S\sim P_0} \   \left[\ell(x;S)\right], \nonumber
\end{flalign}
where the objective function is the expectation of the loss function with respect to the training distribution $P_0$. 
%
To solve the distributional mismatch between training data and testing data, the formulation of DRO \cite{goodfellow2014explaining,madry2017towards,rahimian2019distributionally} was developed, where the goal is to minimize the expected loss with respect to the worst distribution in an uncertainty set $\mathcal U(P_0)$:
\begin{flalign}
    \inf_x \sup_{Q\sim \mathcal U(P_0)} \mathbb E_{S\sim Q} \ \ell(x;S).
\end{flalign}
DRO problems under different uncertainty sets are fundamentally different. Consider the uncertainty set defined by $\varphi$-divergence $D_\varphi(Q\|P_0)$, which is one of the most common choices in the literature and can be written as 
$D_\varphi(Q\|P_0):=\int \varphi\left(\frac{dQ}{dP_0}\right)dP_0,
$
where $\varphi$ is a non-negative convex function such that $\varphi(1)=0$ and $\varphi(t)=+\infty$ for ant $t<0$. 
Then let the uncertainty set $\mathcal U(P_0):=\{Q:D_\varphi(Q\|P_0)\le \rho \}$ where $\rho>0$ is the radius of the uncertainty set.

In this paper, we study the general Cressie-Read family of $\varphi$-divergence \cite{cressie1984multinomial,van2014renyi}, where 
\begin{flalign}
    \varphi_k(t):=\frac{t^k-kt+k-1}{k(k-1)}, \label{define:cressie}
\end{flalign}
$k\in (-\infty,+\infty)\setminus \{0,1\}$. Let $k_*=\frac{k}{k-1}$. This family includes as special cases  $\chi^2$-divergence ($k=2$) and KL divergence ($k\to 1$). When $k>2$, the conjugate function of $\varphi_k(t)$ (which will be introduced later) is not smooth, thus the problem becomes hard to solve even in the penalized formulation \cite{jin2021non}. 
In this paper, we focus on $k\in (1,2]$ ($k_*\in [2,\infty)$). The objective is
\begin{flalign}
\inf_x \sup_{Q:D_{\varphi_k}(Q\|P_0)\le \rho} \mathbb E_{S\sim Q} \ \ell(x;S).\label{eq:objective}
\end{flalign}

Solving \eqref{eq:objective} directly is challenging due to the sup over $Q$. In \cite{namkoong2016stochastic}, a finite-dimensional vector $\textbf{q}$ was used to parameterize the distributions in the uncertainty set since $Q\ll P_0$ for $\varphi$-divergence. Then the DRO problem becomes a convex concave min-max problem. This method can be extended to the case with non-convex loss function by applying the algorithms for non-convex concave min-max problems \cite{rafique2022weakly,lin2020gradient,xu2023unified}. However, the dimension of distribution in the uncertainty set is equal to the number of training samples. Thus, the computational complexity at each iteration is linear in the sample size and is prohibitive in
large-scale applications.

To obtain a complexity independent of the sample size, one alternative is to use its dual. By duality, we can show that the DRO objective \eqref{eq:objective} can be equivalently written as \cite{levy2020large,shapiro2017distributionally}
\begin{flalign}
    \inf_x \inf_{\lambda\ge 0,\tilde\eta \in \mathbb R} \mathbb E_{S\sim P_0}\left[\lambda \varphi_k^*\left(\frac{\ell(x;S)-\tilde\eta}{\lambda}\right)+\lambda\rho+\tilde\eta\right],\nonumber
\end{flalign}
where $\varphi_k^*(t')=\sup_{t}\{t't-\varphi_k(t)\}$ is the conjugate function of $\varphi_k(t')$. In this way, the optimization problem under an unknown distribution is rewritten into one under a known distribution. The subsampling method can then be used, which leads to a complexity independent of the sample size (which will be introduced later).
For the Cressie-Read family in \eqref{define:cressie}, the corresponding conjugate function family is 
$
\varphi_k^*(t)=\frac{1}{k}\left[((k-1)t+1)_+^{k_*} -1\right].
$
Therefore, the objective can be written as 
\begin{flalign}
    \inf_{x,\lambda\ge 0,\tilde\eta \in \mathbb R} &\mathbb E_{ S\sim P_0}\left[ \frac{\lambda}{k}\left((k-1)\frac{\ell(x;S)-\tilde\eta}{\lambda}+1\right)_+^{k_*}\right]\nonumber\\
    &+\lambda\left(\rho-\frac{1}{k}\right)+\tilde\eta\nonumber.
\end{flalign}
Let $\eta=\tilde\eta-\frac{\lambda}{k-1}$ and the corresponding objective is 
\begin{flalign}
\inf_{x,\lambda\ge 0,\eta \in \mathbb R} \mathbb E_{S\sim P_0}&\left[ \frac{(k-1)^{k_*}}{k}({\ell(x;S)-\eta})_+^{k_*}\lambda^{1-k_*} \right]\nonumber\\
&+\lambda\left(\rho+\frac{1}{k(k-1)}\right)+\eta. \nonumber
\end{flalign}
Define
\begin{flalign}
    f(x;\lambda;\eta;s)=& \frac{(k-1)^{k_*}}{k}({\ell(x;S)-\eta})_+^{k_*}\lambda^{1-k_*} \nonumber\\
    &+\lambda\left(\rho+\frac{1}{k(k-1)}\right)+\eta.\label{eq:define1}
\end{flalign}
Thus the goal is to solve
\begin{flalign}
    \inf_x \inf_{\lambda\ge 0,\eta \in \mathbb R} F(x;\lambda;\eta),\label{eq:goal}
\end{flalign}
where $F(x;\lambda;\eta)$ is defined as 
$
    F(x;\lambda;\eta)=\mathbb E_{S\sim P_0}\Big[f(x;\lambda;\eta;S)\Big].
$
Therefore, we reformulate the DRO problem as one to minimize an objective function under a known distribution, where subsampling method could be used to reduce the complexity.
\section{Analysis of Constrained DRO}
In this section, we analyze the constrained DRO problems under Cressie-Read family divergence uncertainty sets with general smooth non-convex loss function. We first discuss the challenges appearing in constrained formulations, then we present how to construct the corresponding approximated problem in order to overcome these challenges.
\subsection{Smooth and Lipschitz Approximation}\label{sec:approx}
For $\lambda\in[0,+\infty),\eta\in \mathbb R$, the objective $F(x;\lambda;\eta)$ is neither smooth nor Lipschitz. Thus it is difficult to implement gradient-based algorithms. In the following, we will construct an approximation of the original problem so that the objective function $F(x;\lambda;\eta)$ becomes smooth and Lipschitz by constraining both $\lambda$ and $\eta$ in some bounded intervals.

Denote by {$\omega=(k(k-1)\rho+1)^{\frac{1}{k}}$}. Since the loss function is bounded such that $0\le \ell\le B$, we can show that there exists an upper bound $\bar \lambda=(k-1){(k(k-1)\rho+1)^{-\frac{1}{k_*}}}\left(1+\frac{(\frac{1}{\omega})^{\frac{1}{k_*-1}}}{1-(\frac{1}{\omega})^{\frac{1}{k_*-1}}}\right)B$ which only depends on $k,\rho$ and $B$ such that the optimal value $\lambda^*\le \bar \lambda$. In this paper, we do not assume that $\lambda^*\geq\lambda_0>0$ as in \cite{wang2021sinkhorn}. Instead, we consider an approximation with $\lambda \in [\lambda_0,\bar \lambda]$, and show that the difference between the orignial and the approximation can be bounded.  
We can show corresponding optimal $\eta^*\in [- \bar \eta,B]$, where $\bar \eta=\bar\lambda\left(\frac{k}{(k-1)^{k_*}k_*}\right)^{\frac{1}{k_*-1}}$. 
The challenge lies in that the value of $\eta$ can be negative. Thus given this $\eta$, the optimal value of $\lambda$ can be quite large then it is hard to  upper bound $\lambda$. In our proof, we change the objective to the function that only depends on $\eta$ and find the lower bound on $\eta$. Based on this lower bound, we  get the bound on $\lambda$.

We show that the difference between the original and the approximation can be bounded in the following lemma. 
\begin{lemma}\label{lemma:duality}
    {$\forall x\in \mathbb R^d, 0\le \lambda_0\le \bar \lambda$,  
\begin{flalign}
    \left|\inf_{\lambda\in [\lambda_0,\bar \lambda],\eta \in [- \bar \eta,B]} F(x;\lambda;\eta)-\inf_{\lambda\ge 0,\eta \in \mathbb R} F(x;\lambda;\eta)\right|\le 2\lambda_0\rho.\nonumber
\end{flalign}
}
\end{lemma}
Note in the proof of this lemma, we {derive an equivalent expression of}
$$\sup_{{Q: }D_{\varphi_k}(Q\|P_0)\le \rho} \mathbb E_{S\sim Q} \ \left[\ell(x;S)\right]-\lambda_0D_{\varphi_k}(Q\|P_0),$$ where both the hard constraint and {regularizer} are kept. This is different from the approach in Section 3.2 of \cite{shapiro2017distributionally}. {Note  {that this equivalent formulation} holds for any $\varphi$-divergence DRO problem.}

Lemma \ref{lemma:duality} demonstrates that the non-smooth objective function can be approximated by a smooth objective function. A smaller $\lambda_0$ makes the gap smaller but the function ``less smooth".
\subsection{Convexity and Smoothness on Parameters}
The advantage of our approximated problem is that the function is smooth in all $x$, $\lambda$, and $\eta$. Moreover, We find that the objective function is convex in $\lambda$ and $\eta$ though the loss function is non-convex in $x$. 
\begin{lemma}\label{lemma:convex_smooth}
    Define $z=(\lambda,
    \eta)\in \mathcal M$, where $\mathcal M=\{(\lambda,\eta):\lambda\in[\lambda_0,\bar \lambda],\eta\in [-\bar\eta, B]\}$. Then  $\forall x\in \mathbb R^d, z\in \mathcal M$, the objective function $F(x;z)$ is convex and $L_z$-smooth in $z$, where $L_z=\frac{1}{\lambda_0}+\frac{2(B+\bar\eta)}{\lambda_0^2}+{\frac{(B+\bar \eta)^2}{\lambda_0^3}}$ if $k_*=2$ and $L_z=\frac{(k-1)^{k_*}}{k}k_*(k_*-1)\left(\frac{(B+\bar \eta)^{k_*}}{\lambda_0^{k_*+1}}+\frac{(B+\bar \eta)^{k_*-2}}{\lambda_0^{k_*-1}}\right)$ if $k_*>2$.

  Moreover, the objective function $F(x;z)$ is $L_x$-smooth in $x$, where $L_x=\frac{(k-1)^{k_*}}{k}k_*\lambda_0^{1-k_*}(B+\bar \eta)^{k_*-2}((k_*-1)G^2+(B+\bar \eta)L)$.
\end{lemma}
Note the first-order gradient of the objective function is non-differential at some point when $k_*=2$. Therefore, we discuss in two cases: $k_*>2$ and $k_*=2$. In the first case, we can get the Hessian matrix of the objective. In the second case, we show the smoothness and convexity.

\begin{algorithm}[tb]
\caption{SFK-DRO}
\label{alg:algorithm}
\textbf{Input}: Iteration number {$T$}, initial point $(x_1,z_1)$, sample numbers $n_x,n_z$, stepsize $\alpha$, and one constant $C$ 
\begin{algorithmic}[1] 
\STATE Let $t=1$ 
\WHILE{$t\le {T}$  }
\STATE randomly select $n_x$ samples {$S_1, S_2, ..., S_{n_x}$} and compute ${\nabla_x f_x(x_t,z_t)}=\sum_{i=1}^{n_x}\frac{\nabla_xf(x_t;z_t;{S_i})}{n_x}$.\\
\STATE $x_{t+1}=x_t-\alpha{\nabla_xf_x}(x_t,z_t)$\\
\STATE randomly select $n_z$ samples {$S_1, S_2, ..., S_{n_z}$}and compute ${\nabla_zf_z}(x_{t+1},z_t)=\sum_{j=1}^{n_z}\frac{\nabla_zf(x_{t+1};z_t;{S_j})}{n_z} $\\
\STATE $e_t$= $\arg\min_{e\in \mathcal M} \langle e,{\nabla_zf_z}(x_{t+1};z_t)\rangle$\\
\STATE $d_t=e_t-z_t$\\
\STATE $g_t=\langle d_t,-{\nabla_zf_z}(x_{t+1};z_t)\rangle$\\
\STATE $\gamma_t=\min\left\{\frac{g_{t}}{C},1\right\}$\\
\STATE $z_{t+1}=z_t+\gamma_td_t$\\
\STATE $t=t+1$\\
\ENDWHILE\\
\end{algorithmic}
$t'=\arg\min_{t}\|{\nabla_xf_x}(x_{t};z_t)\|^2+g_t^2$\\
 \textbf{Output}: $(x_{t'+1},z_{t'})$
\end{algorithm}

\section{Mini-Batch Algorithm}
Existing constrained stochastic {Proximal
Gradient Descent (PGD)} algorithm for general non-convex functions \cite{ghadimi2016mini} can be used to solve the approximated problem directly. {
However, this method requires a small fixed step size, leading to a 4slow convergence. Additionally, it involves a projection step, making it less efficient for large-scale problems.}

{In this paper, we find that the objective $F(x;z)$ is convex in $z$ and non-convex in $x$.}
This motivates us to consider a stronger convergence criterion: 
\begin{flalign}
|\nabla_{x}F(x;\lambda;\eta)|\le \epsilon, \nonumber\\
|F(x;\lambda;\eta)-\inf _{\lambda'\ge 0;\eta'}F(x;\lambda';\eta')|\le \epsilon. \nonumber
\end{flalign}
{In this paper, we use the Frank-Wolfe method to update the variable $z$. Compared with the PGD method, the Frank-Wolfe method can solve the stepsize selection problem by adapting the stepsize through line search techniques and does not require projection, making it efficient for large-scale applications.}
We then provide our  Stochastic gradient and Frank-Wolfe DRO algorithm (SFK-DRO), which optimizes $x$ and $z$ separately (see  Algorithm \ref{alg:algorithm}). Define $D=\max_{z_1,z_2\in \mathcal M}\|z_1-z_2\|$, {$\sigma_0=\frac{(k-1)^{k_*}}{k}k_*(B+\bar \eta)^{k_*-1}G\lambda_0^{1-k_*}$, $\sigma_1=\left(\rho+1+\frac{1}{k(k-1)}\right)+\frac{(k-1)^{k_*}\lambda_0^{-k_*}(B+\bar\eta)^{k_*}}{k}\left(k_*-1+\frac{\lambda_0k_*}{B+\bar \eta}\right)$}, $\Delta=F(x_1;z_1)-\inf_{x,z\in\mathcal M}F(x;z)$ and $C$ is a constant such that $C\ge {D^2}L_z$. The convergence rate is then provided in the following theorem. 
\begin{theorem}\label{theorem:main}
 With mini-batch {sizes} $n_x=\frac{{12L_x\sigma_0^2}}{C\epsilon^2}\sim \mathcal O(\lambda_0^{-2k_*+4}\epsilon^{-2})$, {$n_z\ge \frac{48D^2\sigma_1^2}{\epsilon^2}\sim \mathcal O(\epsilon^{-2k_*-2})$} such that $3B\sqrt{1+k(k-1)\rho}\sqrt{\frac{4+\log(n_z)}{4n_z}}<\frac{\epsilon}{4}$ if $k_*=2$ or  $3B{(1+k(k-1)\rho)^{\frac{1}{k}}}\left(\frac{1}{n_z}+\frac{1}{2^{k_*-1}(k_*-2)n_z}\right)^{\frac{1}{k_*}}<\frac{\epsilon}{4}$ if $k_*>2$
and {$\alpha=\frac{1}{2C}$}, $\lambda_0=\frac{\epsilon}{8\rho}$, for any small $\epsilon>0$ such that $\frac{{D^2}L_z}{L_x}\sim \mathcal O(\epsilon^{-2})\ge2$ and $\frac{{D\sigma_1}}{C}\sim \mathcal O(\epsilon)\le 1$,  at most $T={48}C\Delta\epsilon^{-2}\sim\mathcal O(\lambda_0^{-k_*-1}\epsilon^{-2})$ iterations are needed to guarantee a stationary point $(x_{t'+1};z_{t'})$ in expectation:
\begin{flalign*}
\mathbb E
\|\nabla_{x}F(x_{t'+1};z_{t'})\|&\le \epsilon ,
\\
\mathbb E\Big[\Big|F(x_{t'+1};z_{t'})-\inf _{\lambda\ge0;\eta\in\mathbb R}F(x_{t'+1};\lambda;\eta)\Big|\Big]&\le \epsilon. 
\end{flalign*}
\end{theorem}
A proof sketch will be provided later.  Before that, we introduce a lemma for our subsampling method. Via this lemma, we can show the complexity is independent of the sample size and thus is suitable for our large-scale setting. When we optimize $z$, an estimator ${f_z}(x,z)=\sum_{j=1}^{n_z}\frac{f(x;z;{S_j})}{n_z} $ is build to estimate $F(x;z)=\mathbb E_{S\sim P_0}\Big[f(x;z;S)\Big]$. Though the estimator is unbiased, in our Frank-Wolfe update process \cite{jaggi2013revisiting,frank1956algorithm,lacoste2016convergence} we need to estimate $\min F(x;z)$ via $\mathbb E\min f_z(x;z)$. Obviously, the expectation of minimum is not equal to the minimum of expectation, thus it is a biased estimator. In the following lemma, we show that this gap can be bounded by a decreasing function of the sample batch $n_z$.
\begin{lemma} \label{lemma:bias}
   For any bounded loss function $\ell$, if $k_*=2$,  
\begin{flalign}
    &\left|\inf_{{\lambda\ge0, \eta\in \mathbb R}}\left[F(x_{t+1};\lambda;\zeta)\right]-\mathbb E\left[\inf_{{\lambda\ge0, \eta\in \mathbb R}}f_z(x_{t+1};\lambda;
    \eta)\right]\right|\nonumber\\
    &\le 3B\sqrt{1+k(k-1)\rho}\sqrt{\frac{4+\log(n_z)}{4n_z}};\nonumber
\end{flalign}
and if $k_*>2$, 
\begin{flalign}
    &\left|\inf_{{\lambda\ge0, \eta\in \mathbb R}}\left[F(x_{t+1};\lambda;\eta)\right]-\mathbb E\left[\inf_{{\lambda\ge0, \eta\in \mathbb R}}f_z(x_{t+1};\lambda;\eta)\right]\right|\nonumber\\
    &\le 3B{(1+k(k-1)\rho)^{\frac{1}{k}}}\left(\frac{1}{n_z}+\frac{1}{2^{k_*-1}(k_*-2)n_z}\right)^{\frac{1}{k_*}}.\nonumber
\end{flalign}
\end{lemma}
Note that \cite{levy2020large} only shows this lemma when $k_*=2$, and we extend the results to  $k_*>2$. This lemma shows that the gap is in the order of $\mathcal O({n_z}^{-\frac{1}{k_*}})$ and is independent of the total number of samples. 
\subsection{Proof Sketch of Theorem \ref{theorem:main}}
We use a stochastic gradient descent method \cite{moulines2011non,gower2019sgd,robbins1951stochastic} to update $x$. Since the objective function is $L_x$-smooth in $x$, if $\alpha\le\frac{1}{2L_x}$ we have that:
\begin{flalign}
        \frac{\alpha}{2}&\mathbb E\left[\|\nabla_xF(x_{t};z_t)\|^2\right]\le \mathbb E[F(x_{t};z_t)]-\mathbb E[F(x_{t+1};z_t)]\nonumber\\
    &+{\alpha^2L_x}\mathbb E[\|{\nabla_xf_x}(x_{t};z_t)-\nabla_xF(x_{t};z_t)\|^2],\label{eq:add1}
\end{flalign}
where $f_x(x,z)=\sum_{j=1}^{n_x}\frac{f(x;z;{S_j})}{n_x}$. Define $\sigma_0=\frac{(k-1)^{k_*}}{k}k_*(B+\bar \eta)^{k_*-1}G\lambda_0^{1-k_*}$ and we can show that
\begin{flalign}
   \mathbb E[\|\nabla_xf_x(x_{t};z_t)-\nabla_xF(x_{t};z_t)\|^2]\le \frac{\sigma_0^2}{n_x}. 
\end{flalign}
Since $z\in \mathcal M$, instead of the stochastic gradient descent method, we employ the  Frank-Wolfe method \cite{frank1956algorithm} to update $z$. 
Define $e_t=\arg\min_{e\in \mathcal M} \langle e,{\nabla_zf_z}(x_{t+1};z_t)\rangle$ and $g_t=\langle e_t-z_t,-{\nabla_zf_z}(x_{t+1};z_t)\rangle$. In addition, we have that
\begin{flalign}
    g_t\ge f_z(x_{t+1};z_t)-\min_{z\in \mathcal M}f_z(x_{t+1};z)\nonumber
\end{flalign}
since $f_z(x;z)$ is convex in $z$ \cite{jaggi2013revisiting}. We can show that $\frac{g_t}{C}\sim \mathcal O(\lambda_0)$ thus for small $\lambda_0$ we have $\frac{g_t}{C}\le 1$. 
Then due to the fact that the objective is $L_z$-smooth in $z$ ((9) of \cite{lacoste2016convergence}), we have that 
\begin{flalign}
    \mathbb E\left[\frac{g_t^2}{{4}C}\right]\le \mathbb E[F(x_{t+1};z_t)]-\mathbb E[F(x_{t+1};z_{t+1})]{+\frac{D^2\sigma_1^2}{Cn_z}}.\label{eq:add2}
\end{flalign}
By recursively adding \eqref{eq:add1} and \eqref{eq:add2}, we have that
\begin{flalign}
    &\frac{1}{T}\sum_{t=1}^T\frac{\alpha}{2}\mathbb E\left[\|\nabla_xF(x_{t};z_t)\|^2\right]+\mathbb E\left[\frac{g_t^2}{{4}C}\right]\nonumber\\
    &\le \frac{F(x_{1};z_1)-\mathbb E[F(x_{T+1};z_{T+1})]}{T}+{\alpha^2L_x}\frac{\sigma^2}{n_x}{+\frac{D^2\sigma_1^2}{Cn_z}}.
\end{flalign}
Since $L_z\sim \mathcal O(\lambda_0^{-k_*-1})$ and  $L_z\sim \mathcal O(\lambda_0^{-k_*-1})$, for small $\lambda_0$ we have $C\ge {D^2}L_z\ge 2L_x$. 
Then we set $\alpha=\frac{1}{2C}\le \frac{1}{2L_x}$, $T={48}C\Delta\epsilon^{-2}\sim\mathcal O(\lambda_0^{-k_*-1}\epsilon^{-2}), n_x=\frac{{12}L_x\sigma^2}{C\epsilon^2},$
and denote $\Delta=F(x_1;z_1)-\min_{x,z\in\mathcal M}F(x;z)$,
for some ${t'}\in[1,T]$ we  have that
\begin{flalign}
    \mathbb E\left[\|\nabla_xF(x_{{t'}};z_{t'})\|\right]&\le \frac{\epsilon}{2},\\
    \mathbb E\left[F(x_{{t'}+1};z_{t'})-\inf_{z\in \mathcal M}{ f_z}(x_{{t'}+1};z)\right] \le \mathbb E[g_{t'}]&\le \frac{\epsilon}{2}.
\end{flalign}
We choose $(x_{{t'}+1},z_{t'})$ as our output and we need to bound $\mathbb E\left[\|\nabla_xF(x_{{t'}+1};z_{t'})\|\right]$ and $ \mathbb E\left[F(x_{{t'}+1};z_{t'})-\inf_{z\in \mathcal M}F_z(x_{{t'}+1};z)\right]$.
Since $F(x;z)$ is $L_x$-smooth in $x$, we have that  
$    \mathbb E[\|\nabla_xF(x_{{t'}+1};z_{t'})\|]
    \le \epsilon.$

By Lemma \ref{lemma:bias}, we pick {small} $n_z$ such that $$\left|\inf_{{\lambda\ge 0;\eta\in \mathbb R}}\left[F(x_{{t'}+1};\lambda;\eta)\right]-\mathbb E\left[\inf_{{\lambda\ge 0;\eta\in \mathbb R}}{f_z}(x_{{t'}+1};\lambda;\eta)\right]\right|\le\frac{\epsilon}{4}.$$
{Lemma \ref{lemma:duality} also works for $f_z$}. When $\lambda_0=\frac{\epsilon}{8\rho}$, we have that 
\begin{flalign}
    \left|\inf_{\lambda\in [\lambda_0,\bar \lambda],\eta \in [- \bar \eta,B]} f_z(x;\lambda;\eta)-\inf_{\lambda\ge 0,\eta \in \mathbb R} f_z(x;\lambda;\eta)\right|\le \frac{\epsilon}{4}.\nonumber
\end{flalign}
Thus we have 
\begin{flalign}
    {\mathbb E[}F(x_{{t'}+1};z_{t'})-\inf_{\lambda\ge 0,\eta \in \mathbb R} F(x_{{t'}+1};\lambda;\eta){]}\le \epsilon,
\end{flalign}
which completes the proof.

\section{Smoothed CVaR}
Our algorithm can also solve other DRO problems efficiently, for example, the Smoothed CVaR proposed in \cite{jin2021non}. 
The CVaR DRO is an important $\varphi$-divergence DRO problem, where $\varphi(t)=\mathbb{1}_{[0,\frac{1}{\mu})}$ if $0\le t<\frac{1}{\mu}$, and $0<\mu<1$ is some constant. The dual expression of CVaR can be written as
\begin{flalign}
    \mathcal{L}_{CVaR}(x;P_0)=\inf_{\eta \in \mathbb R} \frac{1}{\mu}\mathbb E_{S\sim P_0}\left[(\ell(x;S)-\eta)_++\eta\right].\nonumber
\end{flalign}
The dual of CVaR is non-differentiable, which is undesirable from an optimization viewpoint. To solve this problem, \cite{jin2021non} proposed a new divergence function, which can be seen as a smoothed version of the CVaR. Their experiment results show the optimization of smoothed CVaR is much easier.  However, \cite{jin2021non}'s method only works for the penalized formulation of DRO. We will show that our method can solve the constrained smoothed CVaR.

Here, the divergence function is 
\begin{flalign}
 \varphi_{s}(t) = \begin{cases}
  t\log(t)+\frac{1-\mu t }{\mu}\log(\frac{1-\mu t}{1-\mu }),  & t \in[0,\frac{1}{\mu}) ;\\
  +\infty, &  \text{ otherwise.}
\end{cases}
\end{flalign}
The corresponding conjugate function is 
\begin{flalign}
    \varphi_{s}^*(t)=\frac{1}{\mu}\log(1-\mu+\mu\exp(t)).
\end{flalign}
The objective function is then written as 
\begin{flalign}
    &\inf_x \inf_{\lambda\ge 0,\eta \in \mathbb R}F_s(x;\lambda;\eta)\nonumber\\
    =& \mathbb E_{S\sim P_0}\left[{\lambda }\varphi_{s}^*(\frac{\ell(x;S)-\eta}{\lambda})+\lambda\rho+\eta\right].
\end{flalign}
We can show that there exist upper bounds for the optimal values $\lambda^*$ and $\eta^*$. There exists a $\bar{\lambda}>0$ only depends on $\mu, B$ and $\rho$ such that $\lambda^*\in [0,\bar \lambda]$ and $\eta^*\in[0,B]$. The proof can be found  in Appendix \ref{appendix:cvar}.

This objective function is non-smooth when $\lambda\to 0$. Therefore, we take a similar approach as the one in \Cref{sec:approx} to approximate the original problem with $\lambda \in [\lambda_0,\bar \lambda]$. We bound the difference in the following lemma. 
\begin{table*}[h!]
\centering
\begin{tabular}{||c c c c c c c c c c c||} 
 \hline
 Class&0&1 &2 & 3 & 4&5&6&7&8&9 \\ [0.5ex] 
 \hline\hline
 EMR & 77.64 & 86.19 & 69.33&54.03&51.53&47.05&87.66&85.35&87.12&83.15 \\ 
 SFK-DRO & 76.11 & 84.71 & 66.18&\textbf{54.95}&\textbf{58.65}&49.36&89.06&84.03&88.41&83.09 \\
 PAN-DRO & 74.92 & 85.62 & 65.72&52.69&55.83&\textbf{49.50}&88.85&84.06&88.68&81.29 \\
 
 \hline
\end{tabular}
\caption{Test Accuracy of each class for imbalanced CIFAR 10.}
\label{table:1}
\end{table*}
\begin{lemma}
    {$\forall x\in \mathbb R^d, \lambda_0\ge0$, 
\begin{flalign}
    \left|\inf_{\lambda\in [\lambda_0,\bar \lambda],\eta \in [0,B]} F_s(x;\lambda;\eta)-\inf_{\lambda\ge 0,\eta \in \mathbb R} F_s(x;\lambda;\eta)\right|\le 2\lambda_0\rho.\nonumber
\end{flalign}
}
\end{lemma}
The proof is similar to Lemma \ref{lemma:duality} thus is omitted here. In addition, we can show that $F_s(x;z)$ is $L_z'$-smooth and convex in $z$, where $L_z'\sim \mathcal O(\lambda_0^{-3})$ if $\lambda\in[\lambda_0,\bar \lambda].$ Also it is easy to get $F_s(x;z)$ is $L_x'$-smooth in $x$, where $L_x'\sim \mathcal O(\lambda_0^{-2})$.

Similar to eq. (42) and Remark 1 in \cite{levy2020large},  we can prove that $\Big|\min_{{\lambda\ge0,\eta\in \mathbb R}}\left[F_s(x_{t+1};\lambda;\eta)\right]-\mathbb E\left[\min_{{\lambda\ge0,\eta\in \mathbb R}}f_s(x_{t+1};\lambda;\eta)\right]\Big|\sim\mathcal O(n_s^{-0.5})$. 
We then use Algorithm 1 directly and the complexity to get {an} $\epsilon$-stationary point is ${\mathcal O(\epsilon^{-9})}$. The detailed proof can be found in  Appendix \ref{appendix:cvar}. 

\section{Numerical Results}
\begin{figure}[!ht]
    \includegraphics[width=\linewidth]{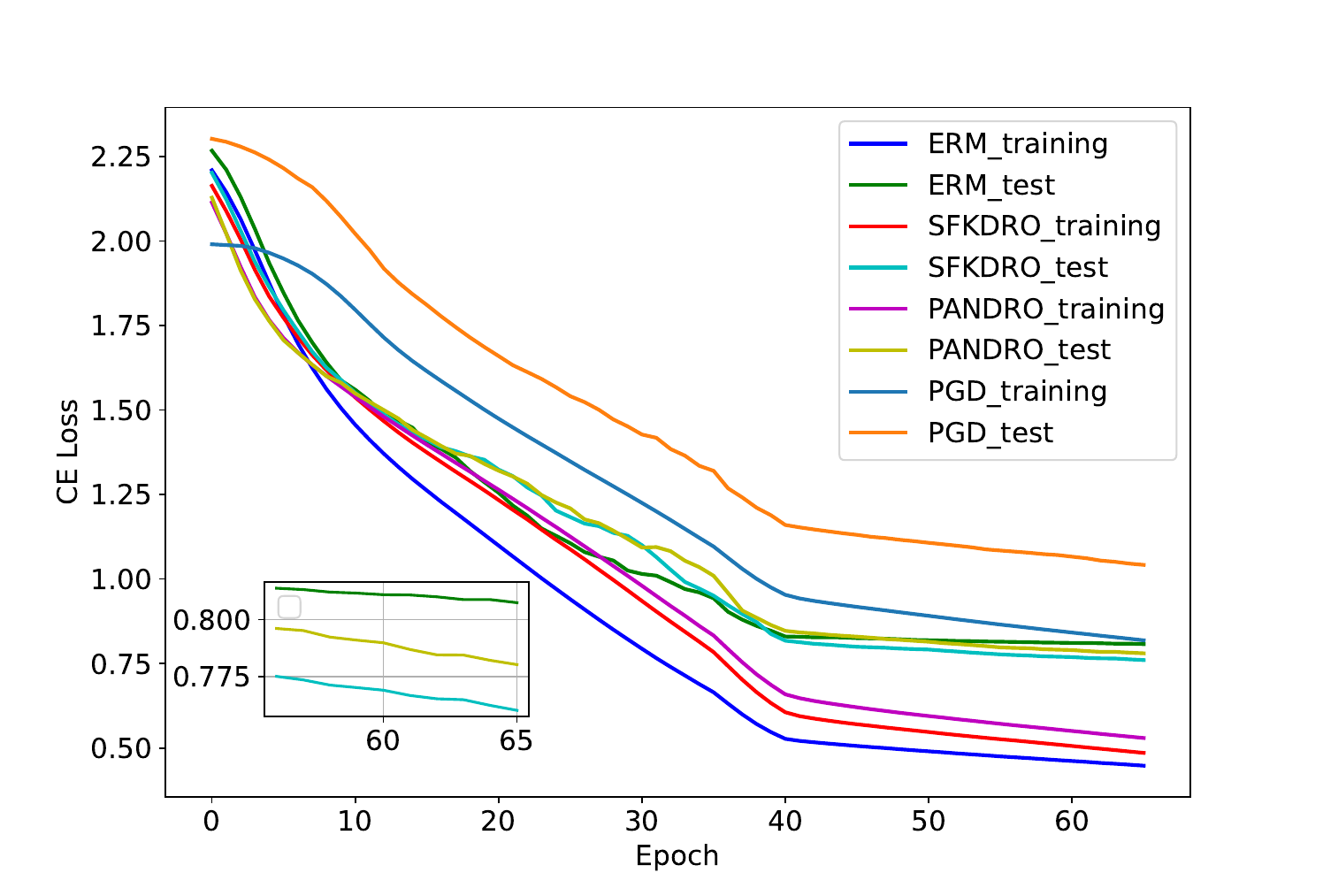}\par\caption{Training curve of classification task.}\label{fig:1}
\vspace{-0.3cm}
\end{figure}

In this section,  we verify our theoretical results in solving an imbalanced classification problem. In the experiment, we consider a non-convex loss function and $k$ is set to be $2$ for the Cressie-Read family. We will show that 1) {although the computational complexity of PGD \cite{ghadimi2016mini} is of the same order as that of our SFK-DRO algorithm, our numerical results indicate that our proposed algorithm achieves faster convergence in practice}; 2) The performance proposed algorithm for the constrained DRO problem outperforms or is close to the performance of the penalized DRO  with respect to the worst classes. Both of them outperform the baseline.\\
\textbf{Tasks.} We conduct experiments on the imbalanced CIFAR-10 dataset, following the experimental setting in \cite{jin2021non,chou2020remix}. The original CIFAR-10 test dataset consists of $10$ classes, where each of the classes has 5000 images. We randomly select training samples from the original set for each class with the following sampling ratio: $\{0.804, 0.543, 0.997, 0.593, 0.390, 0.285, 0.959, 0.806, \\ 0.967, 0.660\}$. We keep the test dataset unchanged.\\
\textbf{Models.} We learn the standard Alexnet model in \cite{krizhevsky2012imagenet} with the standard cross-entropy (CE) loss. For the comparison of convergence rate, we optimize the same dual objective with the PGD algorithm in \cite{ghadimi2016mini}.  To compare robustness, we  optimize the ERM via vanilla SGD. In addition, we propose an algorithm PAN-DRO, which fixes $\lambda$ and only optimizes $\eta$ and the neural network. Thus it gets the solution for the penalized DRO problem.\\
\textbf{Training Details.} We set $\lambda_1=1,\eta_1=0$,   $\lambda_0=0.1, -\bar \eta=-10$, and the upper bounds $\bar \lambda=10,B=10$. To achieve a faster optimization rate, we set the learning rate $\alpha=0.01$ before the first $40$ epochs and $\alpha=0.001$ after. The mini-batch size is chosen to be $128$. All of the results are moving averaged by a window with size $5$. The simulations are repeated by 4 times.\\
\textbf{Results.}
In Figure \ref{fig:1}, we plot the value of the CE loss using different algorithms through the training process. It can be seen that to optimize the same dual objective function with the same learning rate, the PGD algorithm converges slower than our proposed DRO algorithms. Moreover,  compared with ERM, the DRO algorithms have higher training losses but lower test losses, which demonstrates they are robust.

We also provide the test accuracy of trained models in Table \ref{table:1}. It can be shown that for class $3,4,5$, the accuracies are the lowest due to the limited samples. For these classes, the performance of our SFK-DRO algorithm for the constrained DRO is better or close to the performance of PAN-DRO for the penalized DRO. Both DRO algorithms outperform the vanilla ERM algorithm.
\section{Conclusion}
In this paper, we developed the first stochastic algorithm for  large-scale non-convex stochastic constrained DRO problems in the literature with theoretical convergence and complexity guarantee. We developed a smooth and Lipschitz approximation with bounded approximation error to the original problem.Compared with existing algorithms,{  the numerical results show that our proposed algorithm converges faster.} 
The computational complexity at each iteration is independent of the size of the training dataset, and thus our algorithm is applicable to large scale applications. Our results hold for a general family of Cressie-Read divergences. 
\section*{Acknowledgments}
This work of Q. Zhang and S. Zou is supported by the National Science Foundation under Grants CCF-2106560. 
Department of Education. Y. Zhou’s work is supported by the National Science Foundation under Grants CCF-2106216, DMS-2134223 and ECCS-2237830 (CAREER). L. Shen's work is supported by the NSF under Grant DMS-2208385.

This material is based upon work supported under the AI Research Institutes program by National Science Foundation and the Institute of Education Sciences, U.S. Department of Education through Award No. 2229873 - National AI Institute for Exceptional Education.
Any opinions, findings and conclusions or recommendations
expressed in this material are those of the author(s) and do
not necessarily reflect the views of the National Science
Foundation, the Institute of Education Sciences, or the U.S.

\newpage

\bibliography{aaai24}

\appendix


\onecolumn
\appendix

\section{Bounds on the parameters}\label{appendix:bounds}
\begin{proof}
Firstly, we show for bounded loss function $\ell$, the optimal value $\lambda^*$ has an upper bound. If $\lambda^*=0$, then absolutely it has an upper bound. Otherwise, denote by $\lambda^*(\eta)$ for the optimal value of $\lambda$ given $\eta$ and $\lambda^*=\lambda^*(\eta^*)$. We then have $\nabla_\lambda F(x;\lambda^*(\eta);\eta)=0$ since $F(x;\lambda;\eta)$ is convex in $\lambda$ and $\eta$ (which will be shown in ). It then follows that
\begin{flalign}
    \lambda^*(\eta) =(k-1)\omega^{-1}\mathbb E_{S\sim P_0}\left[(\ell(x;S)-\eta)_+^{k_*}\right]^\frac{1}{k_*}.\label{eq:optimal}
\end{flalign}
If $\eta^*\ge 0$, then $\lambda^*(\eta^*)\le(k-1)\omega^{-1}B$. \\
If $\eta^*< 0$, combine \eqref{eq:goal} and \eqref{eq:optimal},
the objective changes into 
\begin{flalign}
    \inf_{x;\eta \in \mathbb R}\bar F(x;\eta)=\omega\left(\mathbb E_{S\sim P_0}(\ell(x;S)-\eta)_+^{k_*}\right)^{1/k_*}+\eta.\nonumber
\end{flalign}
For the optimal value $\eta^*$, we have $\nabla_{\eta} \bar F(x;\eta^*)=0$. It follows that
\begin{flalign}
    &\nabla_{\eta} \bar F(x;\eta^*)\nonumber\\
    =&-\omega\left(\mathbb E_{S\sim P_0}(\ell(x;S)-\eta^*)_+^{k_*}\right)^{\frac{1-k_*}{k_*}}\nonumber\\
    &\times\left(\mathbb E_{S\sim P_0}(\ell(x;S)-\eta^*)_+^{k_*-1}\right)+1=0.\nonumber
\end{flalign}
Therefore, we have 
\begin{flalign}
    \frac{1}{\omega}&=\frac{\mathbb E_{S\sim P_0}(\ell(x;S)-\eta^*)_+^{k_*-1}}{\left[\mathbb E_{S\sim P_0}(\ell(x;S)-\eta^*)_+^{k_*}\right]^{1-\frac{1}{k_*}}}\nonumber\\
    &\ge \frac{|\eta^*|^{k_*-1}}{(B+|\eta^*|)^{k_*-1}},\nonumber 
\end{flalign}
where the last inequality is due to the fact that $\ell(x;S)$ is bounded. Since $\omega>1$, we have that
\begin{flalign}
    |\eta^*|\le \frac{(\frac{1}{\omega})^{\frac{1}{k_*-1}}B}{1-(\frac{1}{\omega})^{\frac{1}{k_*-1}}}.\nonumber
\end{flalign}
Thus, from \eqref{eq:optimal} we have that
\begin{flalign}
    \lambda^*\le (k-1)\omega^{-1}\left(1+\frac{(\frac{1}{\omega})^{\frac{1}{k_*-1}}}{1-(\frac{1}{\omega})^{\frac{1}{k_*-1}}}\right)B=\bar \lambda,
\end{flalign}
where $\bar \lambda$ only depands on the parameter $k$ and the upper bound on the loss function $B$. In addition, for any fixed $\lambda$, for the optimal value $\eta^*(\lambda)$, we have $\nabla_{\eta} F(x;\lambda;\eta^*(\lambda))=0$. Thus, we have 
\begin{flalign}
    \mathbb E_{S\sim P_0}\Big[& ({\ell(x;S)-\eta^*(\lambda)})_+^{k_*-1} \Big]=\lambda^{k_*-1}\frac{k}{(k-1)^{k_*}k_*}.\nonumber
\end{flalign}
Since $\lambda\in [\lambda_0,\bar \lambda],$ we have $\eta^*\in [- \bar \eta,B]$, where $\bar \eta=\bar\lambda\left(\frac{k}{(k-1)^{k_*}k_*}\right)^{\frac{1}{k_*-1}}$. This completes the proof.
\end{proof}
\section{Proof of Lemma \ref{lemma:duality}}\label{appendix:lemma1}
\begin{proof}
We consider the following question first:
$$\sup_{D_{\varphi_k}(Q\|P_0)\le \rho} \mathbb E_{S\sim Q} \ \ell(x;S)-\lambda_0D_{\varphi_k}(Q\|P_0).$$ 
Suppose $\zeta(s)=\frac{dQ(s)}{dP_0(s)}$, then the question can be written as
\begin{flalign}
    \sup_{\zeta\succeq0} \int \ell(x;S)\zeta-\lambda_0\varphi_k(\zeta)dP_0\nonumber\\
    s.t. \int \varphi_k(\zeta)dP_0\le \rho, \int \zeta dP_0=1. \nonumber
\end{flalign}
The Lagrangian of the above problem can be written as 
\begin{flalign}
    \mathcal L(x;\zeta;\lambda;\eta)=&\int \ell(x;S)\zeta-(\lambda_0+\lambda)\varphi_k(\zeta)-\eta\zeta dP_0\nonumber\\
    &+\lambda\rho+\eta.\nonumber
\end{flalign}
The problem is equivalent to
\begin{flalign}
\sup_{\zeta\succeq0} \inf_{\lambda\ge 0,\eta }\mathcal L(x;\zeta;\lambda;\eta).\nonumber
\end{flalign}
Since the Slater condition holds, we can exchange the positions of $\sup$ and $\inf$ thus getting the dual form
\begin{flalign}
    \inf_{\lambda\ge 0,\eta }\lambda\rho+\eta +\sup_{\zeta\succeq0} \int \ell(x;S)\zeta-(\lambda_0+\lambda)\varphi_k(\zeta)-\eta\zeta dP_0.\nonumber
\end{flalign}
Since the maximum operation can be moved inside the integral ( Theorem 14.60 of \cite{rockafellarvariational}), we have that
\begin{flalign}
    &\sup_{\zeta\succeq0} \int \ell(x;S)\zeta-(\lambda_0+\lambda)\varphi_k(\zeta)-\eta\zeta dP_0\nonumber\\  &=\int\sup_{\zeta\succeq0}  \ell(x;S)\zeta-(\lambda_0+\lambda)\varphi_k(\zeta)-\eta\zeta dP_0\nonumber\\
    &=\int\sup_{\zeta\succeq0}  \zeta[\ell(x;S)-\eta]-(\lambda_0+\lambda)\varphi_k(\zeta)dP_0\nonumber\\
    &=\int ((\lambda_0+\lambda)\varphi_k)^*(\ell(x;S)-\eta)dP_0.\nonumber
\end{flalign}
For each  $\lambda>0$ we get $(\lambda \varphi(\ell))^*=\lambda \varphi^*(\frac{\ell}{\lambda})$. Therefore, the objective function changes into 
\begin{flalign}
    \inf_{\lambda\ge 0,\eta }\lambda\rho+\eta +(\lambda_0+\lambda)\mathbb E_{S\sim P_0}\varphi_k^*\left(\frac{\ell(x;S)-\eta}{\lambda_0+\lambda}\right)\nonumber.
\end{flalign}
We then have
\begin{flalign}
    &\inf_{\lambda\in [\lambda_0,\bar \lambda],\eta \in [- \bar \eta,B]} F(x;\lambda;\eta)-\lambda_0\rho\nonumber\\
    &=
    \inf_{\lambda\ge \lambda_0,\tilde\eta \in \mathbb R} \mathbb E_{S\sim P_0}\left[\lambda \varphi_k^*(\frac{\ell(x;S)-\tilde\eta}{\lambda})+(\lambda-\lambda_0)\rho+\tilde\eta\right]\nonumber\\
    &=\sup_{D_{\varphi_k}(Q\|P_0)\le \rho} \mathbb E_{S\sim Q} \ \ell(x;S)-\lambda_0D_{\varphi_k}(Q\|P_0),\label{eq:dual1}
\end{flalign}
where the first equality is due to the definition of $F$, $\lambda^*\in [\lambda_0,\bar \lambda],\eta^* \in [- \bar \eta,B]$, and the second equality is due to the strong duality we provide above. Moreover, we have
\begin{flalign}
    &\sup_{D_{\varphi_k}(Q\|P_0)\le \rho} \mathbb E_{S\sim Q} \ \ell(x;S)\nonumber\\
    &-\sup_{D_{\varphi_k}(Q\|P_0)\le \rho} \mathbb E_{S\sim Q} \ \ell(x;S)-\lambda_0D_{\varphi_k}(Q\|P_0)\nonumber\\
    &\le\lambda_0\rho.\label{eq:dual2}
\end{flalign}
Combining \eqref{eq:goal},\eqref{eq:dual1} and \eqref{eq:dual2}, we complete the proof. 
\end{proof}
\section{Proof of Lemma \ref{lemma:convex_smooth}}
\begin{proof}\label{appendix:lemma2}
From \eqref{eq:define1}, \eqref{eq:goal} we only need to prove $\phi(x;z)=\mathbb E_{S\sim P_0}\left[({\ell(x;S)-\eta})_+^{k_*}\lambda^{1-k_*}\right]$ is convex and smooth in $z$ and smooth in $x$.

Firstly, we have $$\nabla_\lambda\phi(x;z)=(1-k_*)\mathbb E_{S\sim P_0}\left[({\ell(x;S)-\eta})_+^{k_*}\lambda^{-k_*}\right]$$ and $$\nabla_\eta\phi(x;z)=-k_*\mathbb E_{S\sim P_0}\left[({\ell(x;S)-\eta})_+^{k_*-1}\lambda^{1-k_*}\right].$$\\
If $k_*=2$, the problem becomes a $\chi^2$-DRP problem and $\nabla_\eta\phi(z)$ is not differentiable when $\ell(x;S)-\eta=0$. For any $z_1=(\lambda_1;\eta_1),z_2=(\lambda_2;\eta_2)$ where $\lambda_1,\lambda_2\in[\lambda_0,\bar\lambda]$, we have that for any fixed $s\in \mathbb S$ and $a\in[0,1]$
\begin{flalign}
   &2\lambda_1\lambda_2(\ell(x;s)-\eta_1)_+(\ell(x;s)-\eta_2)_+\nonumber\\
   &\le \lambda_1^2(\ell(x;s)-\eta_2)_+^2
   +\lambda_2^2(\ell(x;s)-\eta_1)_+^2.\nonumber
\end{flalign}
Thus, we have 
\begin{flalign}
&\lambda_1\lambda_2\big(a^2(\ell(x;s)-\eta_1)_+^2+(1-a)^2(\ell(x;s)-\eta_2)_+^2
   \nonumber\\
   &+2a(1-a)(\ell(x;s)-\eta_1)_+(\ell(x;s)-\eta_2)_+\big)\nonumber\\
   \le& (a(1-a)\lambda_1^2+(1-a)^2\lambda_1\lambda_2)(\ell(x;s)-\eta_2)_+^2\nonumber\\
   &+(a(1-a)\lambda_2^2+a^2\lambda_1\lambda_2)(\ell(x;s)-\eta_1)_+^2.\label{eq:convex1}
\end{flalign}
In addition, we have
\begin{flalign}
    &\left(\ell(x;s)-{a\eta_1+(1-a)\eta_2}\right)_+^2\nonumber\\
    \le& a^2(\ell(x;s)-\eta_1)_+^2+(1-a)^2(\ell(x;s)-\eta_2)_+^2\nonumber\\
    &+2a(1-a)(\ell(x;s)-\eta_1)_+(\ell(x;s)-\eta_2)_+.\label{eq:convex2}
\end{flalign}
Combine \eqref{eq:convex1} and \eqref{eq:convex2}, we can get 
\begin{flalign}
    &\frac{1}{a\lambda_1+(1-a)\lambda_2}\left(\ell(x;s)-{a\eta_1-(1-a)\eta_2}\right)^2_+\nonumber\\
    \le&\left(\frac{a}{\lambda_1}(\ell(x;s)-\eta_1)_+^2+\frac{1-a}{\lambda_2}(\ell(x;s)-\eta_2)_+^2\right).
\end{flalign}
Take expectations for both sides, we have
\begin{flalign}
    \phi\left(x;az_1+(1-a)z_2\right)\le a\phi(x,z_1)+(1-a)\phi(x,z_2),\nonumber
\end{flalign}
which demonstrates both $\phi(x;z)$ and $F(x;z)$ is convex in $z$. We then show $F$ is smooth in $z$. We have that 
\begin{flalign}
    \|&\nabla_z\phi(x;z_1)-\nabla_z\phi(x;z_2))\|\nonumber\\
    =&\left|2\mathbb E_{s\sim P_0}\left[\frac{(\ell(x;s)-\eta_1)_+}{\lambda_1}-\frac{(\ell(x;s)-\eta_2)_+}{\lambda_2}\right]\right|\nonumber\\
    &+\left|\mathbb E_{s\sim P_0}\left[\frac{(\ell(x;s)-\eta_1)_+^2}{\lambda_1^2}-\frac{(\ell(x;s)-\eta_2)_+^2}{\lambda_2^2}\right]\right|\nonumber\\
    \le &\left|2\mathbb E_{s\sim P_0}\left[\frac{(\ell(x;s)-\eta_1)_+}{\lambda_1}-\frac{(\ell(x;s)-\eta_2)_+}{\lambda_1}\right]\right|\nonumber\\
    &+\left|2\mathbb E_{s\sim P_0}\left[\frac{(\ell(x;s)-\eta_2)_+}{\lambda_1}-\frac{(\ell(x;s)-\eta_2)_+}{\lambda_2}\right]\right|\nonumber\\
    &+\left|\mathbb E_{s\sim P_0}\left[\frac{(\ell(x;s)-\eta_1)_+^2}{\lambda_1^2}-\frac{(\ell(x;s)-\eta_2)_+^2}{\lambda_1^2}\right]\right|\nonumber\\
    &+\left|\mathbb E_{s\sim P_0}\left[\frac{(\ell(x;s)-\eta_2)_+^2}{\lambda_1^2}-\frac{(\ell(x;s)-\eta_2)_+^2}{\lambda_2^2}\right]\right|\nonumber\\
    \le & \frac{2|\eta_1-\eta_2|}{\lambda_0}+\frac{2(B+\bar\eta)}{\lambda_0^2}|\lambda_1-\lambda_2|+\frac{2(B+\bar\eta)}{\lambda_0^2}|\eta_1-\eta_2|\nonumber\\
    &+\frac{(B+\bar \eta)^2}{\lambda_0^3}|\lambda_1-\lambda_2|\nonumber\\
    \le&\left(\frac{2}{\lambda_0}+\frac{4(B+\bar\eta)}{\lambda_0^2}+\frac{{2}(B+\bar \eta)^2}{\lambda_0^3}\right)|z_1-z_2|.\nonumber
\end{flalign}
Therefore, $\phi(x;z)$ is $\frac{2}{\lambda_0}+\frac{4(B+\bar\eta)}{\lambda_0^2}+\frac{{2}(B+\bar \eta)^2}{\lambda_0^3}$-smooth and $F(x,z)$ is $\frac{1}{\lambda_0}+\frac{2(B+\bar\eta)}{\lambda_0^2}+\frac{(B+\bar \eta)^2}{\lambda_0^3}$-smooth in $z$.\\
If $k_*>2$, $\nabla_\eta\phi(z)$ is differentiable. We can get the Hessian matrix of $\phi$ with respect to $z$ as:
\begin{flalign}
H=
    \begin{bmatrix}
k_*(k_*-1)\mathbb E_{S\sim P_0}\left[({\ell(x;S)-\eta})_+^{k_*}\lambda^{-k_*-1}\right], \\
k_*(k_*-1)\mathbb E_{S\sim P_0}\left[({\ell(x;S)-\eta})_+^{k_*-1}\lambda^{-k_*}\right]; \\
k_*(k_*-1)\mathbb E_{S\sim P_0}\left[({\ell(x;S)-\eta})_+^{k_*-1}\lambda^{-k_*}\right], \\
k_*(k_*-1)\mathbb E_{S\sim P_0}\left[({\ell(x;S)-\eta})_+^{k_*-2}\lambda^{1-k_*}\right]
\end{bmatrix}.\nonumber
\end{flalign}
Suppose $a_1,a_2$ are the eigenvalues of $H$. We have
\begin{flalign}
    a_1+a_2&=tr(H)=k_*(k_*-1)\mathbb E_{S\sim P_0}\Big[\nonumber\\
    &({\ell(x;S)-\eta})_+^{k_*}\lambda^{-k_*-1}+({\ell(x;S)-\eta})_+^{k_*-2}\lambda^{1-k_*}\Big]\nonumber\\
    &\ge0\nonumber
\end{flalign}
and
\begin{flalign}
    a_1a_2&=det(H)=k_*^2(k_*-1)^2\lambda^{-2k_*}\nonumber\\
    \times&\Big(\mathbb E_{S\sim P_0}({\ell(x;S)-\eta})_+^{k_*}\mathbb E_{S\sim P_0}({\ell(x;S)-\eta})_+^{k_*-2}\nonumber\\
    -& \left(E_{S\sim P_0}({\ell(x;S)-\eta})_+^{k_*-1}\right)^2 \Big)\nonumber
    \ge 0.
\end{flalign}
Thus $H$ is semi-positive definite which demonstrates $\phi$ is convex in $z$. Moreover, the smooth constant should be the largest eigenvalue.  Therefore we get 
\begin{flalign}
    L_z=\frac{(k-1)^{k_*}}{k}k_*(k_*-1)\left(\frac{(B+\bar \eta)^{k_*}}{\lambda_0^{k_*+1}}+\frac{(B+\bar \eta)^{k_*-2}}{\lambda_0^{k_*-1}}\right)\nonumber
\end{flalign}
Now we prove the objective is $L_x$-smooth in $x$.
Firstly, we have $$\nabla_x\phi(x;z)=k_*\lambda^{1-k_*}\mathbb E_{S\sim P_0}\left[({\ell(x;S)-\eta})_+^{k_*-1}\nabla_x\ell\right].$$
For any $x_1,x_2$ we have that
\begin{flalign}
    \|&\nabla_x\phi(x_1;z)-\nabla_x\phi(x_2;z))\|\nonumber\\
    \le&k_*\lambda^{1-k_*}\big\|\mathbb E_{S\sim P_0} \big[({\ell(x_1;S)-\eta})_+^{k_*-1}\nabla_x\ell(x_1;S)\nonumber\\
    &-({\ell(x_2;S)-\eta})_+^{k_*-1}\nabla_x\ell(x_1;S)\big]\big\|\nonumber\\
    &+k_*\lambda^{1-k_*}\big\|\mathbb E_{S\sim P_0}\big[({\ell(x_2;S)-\eta})_+^{k_*-1}\nonumber\\&\times (\nabla_x\ell(x_2;S)-\nabla_x\ell(x_1;S))\big]\big\|\nonumber.
\end{flalign}
Since we have 
\begin{flalign}
   &\|({\ell(x_1;S)-\eta})_+^{k_*-1}-({\ell(x_2;S)-\eta})_+^{k_*-1}\|\nonumber\\
   \le&(B+\bar \eta)^{k_*-2}\|\ell(x_1;S)-\ell(x_2;S)\|\nonumber\\
   \le& (k_*-1)(B+\bar \eta)^{k_*-2}G\|x_1-x_2\|,\nonumber
\end{flalign}
 where the first inequality is because both $\ell(x;S)$ and $\eta$ are bounded. And the second inequality is due to the fact that $\ell$ is smooth. Thus we have that 
 \begin{flalign}
      &\|({\ell(x_1;S)-\eta})_+^{k_*-1}-({\ell(x_2;S)-\eta})_+^{k_*-1}\|\nonumber\\
   \le&k_*\lambda^{1-k_*}(B+\bar \eta)^{k_*-2}(k_*-1)G^2\|x_1-x_2\|\nonumber\\
   &+k_*\lambda^{1-k_*}(B+\bar \eta)^{k_*-1}L\|x_1-x_2\|.\nonumber
 \end{flalign}
 Therefore, $\phi(x;z)$ is $k_*\lambda^{1-k_*}(B+\bar \eta)^{k_*-2}((k_*-1)G^2+(B+\bar \eta)L)$-smooth and $F(x,z)$ is $\frac{(k-1)^{k_*}}{k}k_*\lambda^{1-k_*}(B+\bar \eta)^{k_*-2}((k_*-1)G^2+(B+\bar \eta)L)$-smooth in $x$.

\end{proof}
\section{Proof of Theorem \ref{theorem:main}}\label{appendix:theorem}
\begin{proof}
    For the update of $x$, we have that
    \begin{flalign}
        x_{t+1}=x_t-\alpha\nabla_xf_x(x_t;z_t).\nonumber
    \end{flalign}  
Since $F(x,z)$ is $L_x$-smooth in $x$, we have
\begin{flalign}
    &F(x_{t+1};z_t)\nonumber\\
    \le&F(x_{t};z_t)+\langle\nabla_xF(x_{t};z_t),x_{t+1}-x_t\rangle\nonumber\\
    &+\frac{L_x}{2}\|x_{t+1}-x_t\|^2\nonumber\\
    =&F(x_{t};z_t)-\alpha\nabla_xf_x(x_{t};z_t)^\top\nabla_xF(x_{t};z_t)\nonumber\\
    &+\frac{\alpha^2L_x}{2}\|\nabla_xf_x(x_{t};z_t)\|^2\nonumber\\
    =&F(x_{t};z_t)-\alpha\nabla_xf_x(x_{t};z_t)^\top\nabla_xF(x_{t};z_t)\nonumber\\
    &+\frac{\alpha^2L_x}{2}\|\nabla_xf_x(x_{t};z_t)-\nabla_x F(x_{t};z_t)+\nabla_x F(x_{t};z_t)\|^2\nonumber\\
    \le&F(x_{t};z_t)-\alpha\nabla_xf_x(x_{t};z_t)^\top\nabla_xF(x_{t};z_t)\nonumber\\
    &+{\alpha^2L_x}\|\nabla_xf_x(x_{t};z_t)-\nabla_x F(x_{t};z_t)\|^2\nonumber\\
    &+{\alpha^2L_x}\|\nabla_x F(x_{t};z_t)\|^2.
\end{flalign}
Given $x_t$ and $z_t$, take the expectation for both sides of (24), we have that
\begin{flalign}
    &\mathbb E[F(x_{t+1};z_t)|x_t,z_t]\nonumber\\
    \le&F(x_{t};z_t)-\alpha\|\nabla_xF_x(x_{t};z_t)\|^2\nonumber\\
    &+{\alpha^2L_x}\mathbb E[\|\nabla_xf_x(x_{t};z_t)-\nabla_xF(x_{t};z_t)\|^2|x_t,z_t]\nonumber\\
    &+{\alpha^2L_x}\|\nabla_xF(x_{t};z_t)\|^2.
\end{flalign}
If $\alpha\le\frac{1}{2L_x}$ we have that $\alpha^2L_x\le \frac{\alpha}{2}$ and
\begin{flalign}
    \frac{\alpha}{2}\|\nabla_xF_x(x_{t};z_t)\|^2\le& F(x_{t};z_t)-\mathbb E[F(x_{t+1};z_t)|x_t,z_t]\nonumber\\
    &+{\alpha^2L_x}\frac{\sigma_0^2}{n_x},\nonumber
\end{flalign}
where the inequality is because $\nabla_xf(x;z;s) \le \frac{(k-1)^{k_*}}{k}k_*(B+\bar \eta)^{k_*-1}G\lambda_0^{1-k_*}=\sigma_0$ is bounded. After that, we take expectations for both sides and we have 
\begin{flalign}
        \frac{\alpha}{2}\mathbb E\left[\|\nabla_xF_x(x_{t};z_t)\|^2\right]\le& \mathbb E[F(x_{t};z_t)]-\mathbb E[F(x_{t+1};z_t)]\nonumber\\
    &+{\alpha^2L_x}\frac{\sigma_0^2}{n_x}.\label{eq:add3}
\end{flalign}
For the update of $z$ and $\forall \gamma_t\in[0,1]$, we get an affine invariant version of the standard descent Lemma ((1.2.5) in \cite{nesterov2003introductory})
\begin{flalign}
    &F(x_{t+1};z_{t+1})\nonumber\\
    \le&F(x_{t+1};z_t)+\gamma_t\langle\nabla_z f_z(x_{t+1};z_t),d_t\rangle+\gamma_t\langle \nabla_z F(x_{t+1};z_t)-\nabla_z f_z(x_{t+1};z_t),d_t\rangle+\frac{\gamma_t^2}{2}C,\nonumber
\end{flalign}
where $C\ge D^2L_z$. In our algorithm we have $\gamma_t=\min\left\{\frac{g_t}{C},1\right\}$ and 
\begin{flalign}
    \frac{g_t}{C}&\le \frac{D\|\nabla_zf_z(x_{t+1};z_t)\|}{D^2L_z}\nonumber\\
    &\le\frac{(\rho+\frac{1}{k(k-1)})+\frac{(k-1)^{k_*}(k_*-1)}{k}\lambda_0^{-k_*}}{DL_z}.\nonumber
\end{flalign}
Since $L_z\sim \mathcal O(\lambda_0^{-k_*-1})$ thus for small $\lambda_0$ we have $\frac{g_t}{C}\le 1$. Consequently, we can assume $\gamma=\frac{g_t}{C}$ and we have 
\begin{flalign}
&F(x_{t+1};z_{t+1})\nonumber\\
    \le&F(x_{t+1};z_t)-\frac{g_t}{C}g_t+\frac{g_t^2}{4C}+\frac{D^2\|\nabla_z F(x_{t+1};z_t)-\nabla_z f_z(x_{t+1};z_t)\|^2}{C}+\frac{(\frac{g_t}{C})^2}{2}C.   \label{eq:fk1}
\end{flalign}
Since $f_z(x;z)$ is convex in $z$, we have that $g_t\ge f_z(x_{t+1};z_t)-\min_{z\in \mathcal M}f_z(x_{t+1};z).$ Take expectations for both sides of \eqref{eq:fk1}, we have that
\begin{flalign}
    \mathbb E\left[\frac{g_t^2}{4C}\right]\le \mathbb E[F(x_{t+1};z_t)]-\mathbb E[F(x_{t+1};z_{t+1})]+\frac{D^2\sigma_1^2}{Cn_z}.\label{eq:add4}
\end{flalign}
By recursively adding \eqref{eq:add3} and \eqref{eq:add4}, we have that
\begin{flalign}
    &\frac{1}{T}\sum_{t=1}^T\frac{\alpha}{2}\mathbb E\left[\|\nabla_xF_x(x_{t};z_t)\|^2\right]+\mathbb E\left[\frac{g_t^2}{4C}\right]\nonumber\\
    &\le \frac{F(x_{1};z_1)-\mathbb E[F(x_{T+1};z_{T+1})]}{T}+{\alpha^2L_x}\frac{\sigma^2}{n_x}+\frac{D^2\sigma_1^2}{Cn_z}.\nonumber
\end{flalign}
Since $D^2L_z\sim \mathcal O(\lambda_0^{-k_*-1})$ and $L_x\sim \mathcal O(\lambda_0^{-k_*+1})$, we can find $\lambda_0$ small enough such that $C\ge D^2L_z\ge 2L_x$. Set $\alpha=\frac{1}{2C}$, we then have that
\begin{flalign}
    \frac{1}{T}\sum_{t=1}^T\mathbb E\left[\|\nabla_xF_x(x_{t};z_t)\|^2\right]+\mathbb E\left[{g_t^2}\right]\le \frac{4C\Delta}{T}+\frac{L_x\sigma^2}{n_xC}+\frac{4D^2\sigma_1^2}{n_z}.\nonumber
\end{flalign}
From Jensen's inequality, we have that
\begin{flalign}
    \frac{1}{T}\sum_{t=1}^T\mathbb E\left[\|\nabla_xF_x(x_{t};z_t)\|\right]^2+\mathbb E\left[{g_t}\right]^2\le \frac{2C\Delta}{T}+\frac{L_x\sigma^2}{n_xC}+\frac{4D^2\sigma_1^2}{n_z}.\nonumber
\end{flalign}
When we set $T=48C\Delta\epsilon^{-2}\sim\mathcal O(\lambda_0^{-k_*-1}\epsilon^{-2}), n_x=\frac{12L_x\sigma^2}{C\epsilon^2},$
for some $t'\in[1,T]$ we  have $$\mathbb E\left[\|\nabla_xF(x_{t'};z_{t'})\|\right]\le \frac{\epsilon}{2}$$
and 
$$\mathbb E[g_{t'}]\le \frac{\epsilon}{2}$$
Since $F(x;z)$ is $L_x$-smooth in $x$, we have that 
\begin{flalign}
    \|\nabla_xF(x_{t'+1};z_{t'})-\nabla_xF(x_{t'};z_{t'})\|
    &\le L_x\|x_{t'+1}-x_{t'}\|\nonumber\\
    &\le L_x\alpha  \|\nabla_xf_x(x_{t'};z_{t'})\|\nonumber.
\end{flalign}
In addition,
\begin{flalign}
    &\mathbb E[\|\nabla_xf_x(x_{t'};z_{t'})\|^2]\nonumber\\
    \le& 
    2\mathbb E[\|\nabla_xf_x(x_{t'};z_{t'})-\nabla_xF(x_{t'};z_{t'})\|^2|x_{t'},z_{t'}]\nonumber\\
    &+2\mathbb E[\|\nabla_xF(x_{t'};z_{t'})\|^2]\nonumber\\
    \le &\frac{\epsilon^2}{2}+\frac{C\epsilon^2}{4L_x}\nonumber
\end{flalign}
Thus 
\begin{flalign}
    \mathbb E\left[L_x\alpha  \|\nabla_x f_x(x_{t'};z_{t'})\|\right]\le \frac{L_x}{C}\sqrt{\frac{\epsilon^2}{2}+\frac{C\epsilon^2}{4L_x}}\le \frac{\epsilon}{2}\nonumber
\end{flalign}
since $C\ge 2L_x$.
Therefore, we have
\begin{flalign}
    \mathbb E[\|\nabla_xF(x_{t'+1};z_{t'})\|]\le  \epsilon.\nonumber
\end{flalign}
In addition, we have
\begin{flalign}
    g_{t'}\ge f_z(x_{t'+1};z_{t'})-\min_{z\in \mathcal M}f_z(x_{t'+1};z).\nonumber
\end{flalign}
Given $x_{t'+1}$ and $z_{t'}$, take expectations for both sides and we have
\begin{flalign}
    \mathbb E[g_{t'}|x_{t'+1},z_{t'}]\ge F(x_{t'+1};z_{t'})-\mathbb E\left[\min_{z\in \mathcal M}f_z(x_{t'+1};z)\right].\nonumber
\end{flalign}
Lemma \ref{lemma:duality} also works for $\widehat F_z$. By Lemma \ref{lemma:duality}, when $\lambda_0=\frac{\epsilon}{8\rho}$, we have 
\begin{flalign}
    \left|\inf_{\lambda\in [\lambda_0,\bar \lambda],\eta \in [- \bar \eta,B]} f_z(x;\lambda;\eta)-\inf_{\lambda\ge 0,\eta \in \mathbb R} f_z(x;\lambda;\eta)\right|\le \frac{\epsilon}{4}.\nonumber
\end{flalign}
By Lemma \ref{lemma:bias} we can get a small $n_z$ such that $3B\sqrt{1+k(k-1)\rho}\sqrt{\frac{4+\log(n_z)}{4n_z}}<\frac{\epsilon}{4}$ if $k_*=2$ or  $3B{(1+k(k-1)\rho)^{\frac{1}{k}}}\left(\frac{1}{n_z}+\frac{1}{2^{k_*-1}(k_*-2)n_z}\right)^{\frac{1}{k_*}}<\frac{\epsilon}{4}$ if $k_*>2$.  We then have that
$$\left|\inf_{\lambda\ge0, \eta\in \mathbb R}\left[F(x_{t'+1};\lambda;\eta)\right]-\mathbb E\left[\inf_{\lambda\ge0, \eta\in \mathbb R}f_z(x_{t'+1};\lambda;\eta)\right]\right|\le\frac{\epsilon}{4}.$$

Thus we have 
\begin{flalign}
    \mathbb E\left[F(x_{t'+1};z_{t'})-\inf_{\lambda\ge 0,\eta \in \mathbb R} F(x_{t'+1};\lambda;\eta)\right]\le \epsilon.
\end{flalign}
which completes the proof.
\end{proof}

\section{Proof of Lemma \ref{lemma:bias}}\label{appendix:lemma3}
The (20) of \cite{levy2020large} provides an inverse-cdf formulation of the DRO problem. By implementing the inverse-cdf formulation, the (42) and remark 1 of \cite{levy2020large} show that 
\begin{flalign}
    &\left|\min_{\lambda\ge0, \eta\in \mathbb R}\left[F(x_{};\lambda;\eta)\right]-\mathbb E\left[\min_{\lambda\ge0, \eta\in \mathbb R}\widehat F_z(x_{};\lambda;\eta)\right]\right|\nonumber\\
    &\le \int_0^1(r(\beta)-r(1))(\beta\cdot h(\beta))'d\beta\nonumber\\
    &\le \|r\|_k\|(\beta\cdot h(\beta))'\|_{k_*},\label{eq:err0}
\end{flalign}
where $r\in \mathcal{R}:=\{r:[0,1]\to \mathbb R_+|\int_0^1r(\beta)d\beta=1 \text{ and } \int_0^1\varphi_k(r(\beta))d\beta\le \rho\}$, $h=3B\min\left\{\sqrt{\frac{1}{\beta n_z}},1\right\}$ and the second inequality is due to the Hölder’s inequality. Note this inequality holds for any fixed $x$, no matter whether the loss function is convex or not. \\
Since $\int_0^1\varphi_k(r(\beta))d\beta\le \rho$, we have that
\begin{flalign}
    \|r\|_k^k\le 1+k(k-1)\rho.\label{eq:err1}
\end{flalign}
Moreover, we have that
\begin{flalign}
   &\|(\beta\cdot h(\beta))'\|_{k_*}^{k_*}\nonumber\\
   &=\int_0^{\frac{1}{n_z}}(3B)^{k_*}d\beta+\int_{\frac{1}{n_z}}^1(\frac{3B}{2})^{k_*}\sqrt{\frac{1}{(\beta n_z)^{k_*}}}d\beta.
\end{flalign}
For $k_*=2$, we have 
\begin{flalign}
    \|(\beta\cdot h(\beta))'\|_2^2=(3B)^2\frac{4+\log(n_z)}{4n_z}\label{eq:err2}
\end{flalign}
and if $k_*>2$, we have that
\begin{flalign}
    \|(\beta\cdot h(\beta))'\|_{k_*}^{k_*}&\le(3B)^{k_*}\left(\frac{1}{n_z}+\frac{1}{2^{k_*}}\frac{2}{{k_*}{}-2}\frac{n^{0.5k_*-1}-1}{n^{0.5k_*}}\right)\nonumber\\
    &\le(3B)^{k_*}\left(1+\frac{1}{2^{k_*-1}(k_*-2)}\right)\frac{1}{n_z}\label{eq:err3}.
\end{flalign}
Combine \eqref{eq:err0},\eqref{eq:err1},\eqref{eq:err2} and \eqref{eq:err3}, we can get the lemma and complete the proof.
\section{Proof of smoothed CVaR}\label{appendix:cvar}
\begin{proof}
\textbf{Bounded parameters:}
We have 
\begin{flalign}
    \nabla_t\varphi_{s}^*(t)=\frac{1}{\mu}\frac{\mu\exp(t)}{1-\mu+\mu\exp(t)}\le \frac{1}{\mu}\nonumber
\end{flalign}
and 
\begin{flalign}
    \nabla_t^2\varphi_{s}^*(t)=\frac{1}{\mu}\frac{\mu(1-\mu)\exp(t)}{(1-\mu+\mu\exp(t))^2}\le \frac{1}{4\mu}\nonumber
\end{flalign}
\begin{flalign}
  \nabla_\eta F_s(x;\lambda;\eta)=  1-\mathbb E_{S\sim P_0}{\varphi_{s}^*}'(\frac{\ell(x;S)-\eta}{\lambda}).\nonumber
\end{flalign}
\begin{flalign}
  \nabla_\eta^2 F_s(x;\lambda;\eta)=  \frac{1}{\lambda}\mathbb E_{S\sim P_0}{\varphi_{s}^*}''(\frac{\ell(x;S)-\eta}{\lambda}).\nonumber
\end{flalign}

If $\eta>B$, then $\nabla_\eta F_s(x;\lambda;\eta)<0$. If $\eta<0$, then $\nabla_\eta F_s(x;\lambda;\eta)>0$. Thus 
$\eta^*\in[0,B]$.

\begin{flalign}
  \nabla_\lambda F_s(x;\lambda;\eta)=  &\rho+\mathbb E_{S\sim P_0}\Big[{\varphi_{s}^*}(\frac{\ell(x;S)-\eta}{\lambda})\nonumber\\
  &-{\varphi_{s}^*}'(\frac{\ell(x;S)-\eta}{\lambda})\frac{\ell(x;S)-\eta}{\lambda}\Big].\nonumber
\end{flalign}
\begin{flalign}
  \nabla_\lambda^2 F_s(x;\lambda;\eta)=\frac{1}{\lambda^3}\mathbb E_{S\sim P_0}\left[{\varphi_{s}^*}''(\frac{\ell(x;S)-\eta}{\lambda})(\ell(x;S)-\eta)^2\right].\nonumber
\end{flalign}
The second-order demonstrates that the $F_s(x;\lambda;\eta)$ is convex in $\lambda$. We then show that $\lambda^*$ has an upper bound. If $\nabla_{\lambda}F_s(x;\lambda;\eta)\ge0$ when $\lambda\to 0$, then $\lambda^*=0.$ If $\nabla_{\lambda}F_s(x;\lambda;\eta)<0$ when $\lambda\to 0$, since both ${\varphi_{s}^*}(t),{\varphi_{s}^*}'(t)$ are increaing with $t$, we have that
\begin{flalign}
\nabla_\lambda F_s(x;\lambda;\eta)> \rho+{\varphi_{s}^*}(\frac{-B}{\lambda})-\frac{B}{\mu\lambda}.
\end{flalign}
Since we know $g(\lambda)=\rho+{\varphi_{s}^*}(\frac{-B}{\lambda})-\frac{B}{\mu\lambda}$ is increasing with $\lambda$ and $g(\lambda)<0$ when $\lambda\to 0$, $g(\lambda)=\rho>0$ when $\lambda\to \infty.$ Thus we can find $\bar \lambda>0$ that $g(\bar \lambda)=0$. Moreover, the value of $\bar \lambda$ only depends on $\mu, B$ and $\rho$.\\
\textbf{Smoothness:}
fix $x$, the Hessian matrix of $F_s(x;z)$ with respect to $z$ as:
\begin{flalign}
H_s=
    \begin{bmatrix}
\frac{1}{\lambda}\mathbb E_{S\sim P_0}{\varphi_{s}^*}''(\frac{\ell(x;S)-\eta}{\lambda}), \\
\frac{1}{\lambda^2}\mathbb E_{S\sim P_0}\left[{\varphi_{s}^*}''(\frac{\ell(x;S)-\eta}{\lambda})(\ell(x;S)-\eta)\right]; \\
\frac{1}{\lambda^2}\mathbb E_{S\sim P_0}\left[{\varphi_{s}^*}''(\frac{\ell(x;S)-\eta}{\lambda})(\ell(x;S)-\eta)\right], \\
\frac{1}{\lambda^3}\mathbb E_{S\sim P_0}\left[{\varphi_{s}^*}''(\frac{\ell(x;S)-\eta}{\lambda})(\ell(x;S)-\eta)^2\right]
\end{bmatrix}.\nonumber
\end{flalign}
Suppose $a_3,a_4$ are the eigenvalues of $H_s$. We have $a_3+a_4>0$ and $a_3a_4\ge 0$. And the function is $L_z'$-smooth and convex in $z$, where $L_z'\sim \mathcal O(\lambda_0^{-3})$ if $\lambda\in[\lambda_0,\bar \lambda].$ Also it is easy to get $F_s(x;\lambda;\eta)$ is $L_x'$-smooth in $x$, where $L_x'\sim \mathcal O(\lambda_0^{-1})$.\\
\textbf{Bounded gap:} denote $f_s(x,z)=\sum_{i=1}^{n_s}\frac{f(x;z;S_i)}{n_s}$, in order to use algorithm 1 directly, we need to estimate $\min F(x;z)$ via $\mathbb E\min f_z(x;z)$ and bound the gap.  From the (42) and remark 1 of \cite{levy2020large},  we have that 
\begin{flalign}
    &\left|\min_{z\in \mathcal M}\left[F_s(x_{t+1};z)\right]-\mathbb E\left[\min_{z\in \mathcal M}f_s(x_{t+1};z)\right]\right|\nonumber\\
    &\le \int_0^1(r(\beta)-r(1))(\beta\cdot h(\beta))'d\beta
\end{flalign}
where $r\in \mathcal{R}:=\{r:[0,1]\to \mathbb R_+|\int_0^1r(\beta)d\beta=1 \text{ and } \int_0^1\varphi_s(r(\beta))d\beta\le \rho\}$, $h=3B\min\left\{\sqrt{\frac{1}{\beta n_s}},1\right\}$  Since $\int_0^1\varphi_s(r(\beta))d\beta\le \rho\}$, we have that $r(\beta)\le \frac{1}{\mu}$ for any 
Moreover, we have that
\begin{flalign}
   &\int_0^1(\beta\cdot h(\beta))'d\beta\nonumber   =\int_0^{\frac{1}{n_s}}3Bd\beta+3B\int_{\frac{1}{n_s}}^1\sqrt{\frac{1}{\beta n_s{}}}d\beta\nonumber\\
   &=\frac{3B}{n_s}+6B(\sqrt{\frac{1}{n_s}}-\frac{1}{n_s})\nonumber
\end{flalign}
Thus the gap $\Big|\min_{\lambda\ge0, \eta\in \mathbb R}\left[F_s(x_{t+1};\lambda;
\eta)\right]-\mathbb E\left[\min_{\lambda\ge0, \eta\in \mathbb R}f_s(x_{t+1};\lambda;\eta)\right]\Big|\sim\mathcal O(n_s^{-0.5})$. 
We then can use algorithm 1 directly and the complexity to get the $\epsilon$-stationary point is $\mathcal O(\epsilon^{-9})$.
\end{proof}
\section{Complexity of PGD}
\subsection{Cressie-Read family}
In the PGD algorithm \cite{ghadimi2016mini}, $y=(x;\lambda;\eta)$ is optimized as  a whole. From Lemma \ref{lemma:convex_smooth}, we know $F(y)=F(x;z)$ is $L_z$-smooth in $z$ and $L_x$-smooth in x. Moreover, it is not hard to show that $\nabla_xF(x;z)$ is $L_{xz}$-Lipschitz in $z$ and $\nabla_zF(x;z)$ is $L_{xz}$-Lipschitz in $x$, where $L_{zx}, L_{xz}\sim\mathcal O(\lambda_0^{-k_*})$.
Therefore, we have 
\begin{flalign}
    \|\nabla_yF(y_1)-\nabla_yF(y_2)\|=&\|\nabla_xF(x_1;z_1)-\nabla_xF(x_2;z_2)\|+\|\nabla_zF(x_1;z_1)-\nabla_zF(x_2;z_2)\|\nonumber\\
    \le& \|\nabla_xF(x_1;z_1)-\nabla_xF(x_1;z_2)\|+\|\nabla_zF(x_1;z_1)-\nabla_zF(x_1;z_2)\|\nonumber\\
    &+\|\nabla_xF(x_1;z_2)-\nabla_xF(x_2;z_2)\|+\|\nabla_zF(x_1;z_2)-\nabla_zF(x_2;z_2)\|\nonumber\\
    \le&L_{xz}\|z_1-z_2\|+L_{z}\|z_1-z_2\|+L_{zx}\|x_1-x_2\|+L_{x}\|x_1-x_2\|\nonumber\\
    \le &(L_x+L_z+L_{xz}+L_{zx})\|y_1-y_2\|.\nonumber
\end{flalign}
Thus, $F(y)$ is $L_y$-smooth in $y$, where $L_y=L_x+L_z+L_{xz}+L_{zx}\sim \mathcal O(\lambda_0^{-k_*-1}).$ According to Corollary 3 in \cite{ghadimi2016mini} and $\lambda_0\sim \mathcal O(\epsilon)$, we can get the $\epsilon$- stationary point with the number of iterations $T\sim \mathcal{O}(\lambda_0^{-k_*-1}\epsilon^{-2})$ and batch size $n_p\sim \mathcal O(\lambda^{-2k_*}\epsilon^{-2}).$ Thus, the total complexity is $\mathcal{O}(\epsilon^{-3k_*-5}).$
\subsection{Smoothed CVaR}
Similar to Cressie-Read family, we can show that $F_s(y)$ is $L_y'$-smooth in $y$, where $L_y'\sim \mathcal O(\lambda_0^{-3}).$ According to Corollary 3 in \cite{ghadimi2016mini}, we can get the $\epsilon$- stationary point with the number of iterations $T\sim \mathcal{O}(\lambda_0^{-3}\epsilon^{-2})$ and batch size $n_p\sim \mathcal O(\lambda^{-2}\epsilon^{-2}).$ Thus, the total complexity is $\mathcal{O}(\epsilon^{-9}).$
\bigskip
\noindent 

\end{document}